\documentclass[letterpaper, 10 pt, conference]{ieeeconf}  

\IEEEoverridecommandlockouts
\usepackage{url}
\usepackage{balance}
\usepackage{amsmath}
\usepackage{amssymb,bm}
\usepackage{amsfonts}
\usepackage{enumerate}
\usepackage{tabulary}
\usepackage{multirow}
\usepackage{cite}
\usepackage{lipsum}  
\usepackage{graphicx}
\usepackage{epstopdf}
\usepackage[misc]{ifsym}
\usepackage{color}
\usepackage[usenames,dvipsnames,table]{xcolor}
\usepackage{xxcolor}
\usepackage{pgf}
\usepackage{tikz}
\usepackage{float}
\usepackage{lipsum}
\usepackage{booktabs}
\usepackage{verbatim}
\usepackage{hyperref}
\usepackage{makecell}
\usepackage{xfrac}
\usepackage[normalem]{ulem}

\usepackage{ifthen}
\usepackage{forloop}
\usepackage{listings}
\usepackage[lined,boxed,commentsnumbered,linesnumbered]{algorithm2e}

\definecolor{codegreen}{rgb}{0,0.6,0}
\definecolor{codepurple}{rgb}{0.58,0,0.82}
\definecolor{backcolour}{rgb}{0.95,0.95,0.92}
\lstdefinestyle{buzz}{
    backgroundcolor=\color{black!5},   
    commentstyle=\color{codegreen},
    keywordstyle=\color{blue},
    numberstyle=\tiny\color{black!30},
    stringstyle=\color{codepurple},
    basicstyle=\footnotesize\ttfamily,
    breakatwhitespace=false,         
    breaklines=true,                 
    captionpos=b,                    
    keepspaces=true,                 
    numbers=left,                    
    numbersep=5pt,                  
    showspaces=false,                
    showstringspaces=false,
    showtabs=false,                  
    tabsize=2,
}
\newcommand{\twodots}{\mathinner {\ldotp \ldotp}}

\SetAlCapSkip{0.5em}

\newcommand{\revision}[1]{{\color{black} #1}}

\newcommand\soutpars[1]{\let\helpcmd\sout\parhelp#1\par\relax\relax}
\long\def\parhelp#1\par#2\relax{\helpcmd{#1}\ifx\relax#2\else\par\parhelp#2\relax\fi }

\newcommand{\sysstate}{\mathbf{x}}
\newcommand{\sysinput}{\mathbf{u}}

\title{\LARGE \bf
\texttt{safe-control-gym}: a Unified Benchmark Suite for Safe Learning-based Control and Reinforcement Learning \revision{in Robotics}
}

\author{
Zhaocong Yuan,
Adam W. Hall,
Siqi Zhou,
Lukas Brunke,
Melissa Greeff,
Jacopo Panerati, Angela P. Schoellig
\thanks{The authors are with the \href{http://www.dynsyslab.org}{Dynamic Systems Lab (DSL)}, Institute for Aerospace Studies, University of Toronto;
the University of Toronto Robotics Institute;
and affiliated with the \href{https://vectorinstitute.ai/}{Vector Institute for Artificial Intelligence} in Toronto. E-mails:
        {\tt \{firstname.lastname\}@robotics.utias.utoronto.ca}}}

\begin{document}
\maketitle
\thispagestyle{empty}
\pagestyle{empty}
\bstctlcite{IEEEexample:BSTcontrol}
\begin{abstract}
In recent years, both reinforcement learning and learning-based control---as well as the study of their \emph{safety}, which is crucial for deployment in real-world robots---have gained significant traction.
However, to adequately gauge the progress and applicability of new results, we need the tools to equitably compare the approaches proposed by the controls and reinforcement learning communities.
Here, we propose a new open-source benchmark suite, called \texttt{\small safe-control-gym}, supporting both model-based and data-based control techniques.
We provide implementations for three dynamic systems---the cart-pole, the 1D, and 2D quadrotor---and two control tasks---stabilization and trajectory tracking.
We propose to extend OpenAI's \emph{Gym} API---the \emph{de facto} standard in reinforcement learning research---with \emph{(i)} the ability to specify (and query) symbolic dynamics and \emph{(ii)} constraints, and \emph{(iii)} (repeatably) inject simulated disturbances in the control inputs, state measurements, and inertial properties.
To demonstrate our proposal and in an attempt to bring research communities closer together, we show how to use \texttt{\small safe-control-gym} to quantitatively compare the control performance, data efficiency, and safety of multiple approaches from the fields of traditional control, learning-based control, and reinforcement learning.
\end{abstract}

\section*{Code Repository}
The open-source code repository of this project is at: {\small \url{https://github.com/utiasDSL/safe-control-gym}}.

\section{Introduction}
\label{sec:intro}

In a near future, robots will be the backbone of transportation, warehousing, and manufacturing. However, for ubiquitous robotics to materialize, we need to devise methods to develop robotic controllers faster and for increasingly complex systems---leveraging data and machine learning to scale up current design approaches.
Top computing hardware and software companies (including  Nvidia~\cite{makoviychuk2021isaac}, Google~\cite{freeman2021brax}, and intrinsic~\cite{tan-white_2021}) are currently working on fast physics-based simulation to complement real-world data and accelerate the progress of robot learning.
At the same time, because safety is a crucial component of cyber-physical systems operating in the real world, safe learning-based control and safe reinforcement learning (RL) have become bustling areas of academic research over the past few years~\cite{DSL2021}.

While the continuous influx of new contributions in safe robot learning reflects the relevance and progress of the field, 
we need to establish ways to fairly compare results yielded by controllers that leverage very different methodologies to understand their advantages and disadvantages. 
This is particularly important as different safe control approaches make different assumptions about the availability of prior knowledge of the system and of experimental data.
Furthermore, we need quantitative benchmarks to assess these controllers' safety and robustness.

\begin{figure}[t]
  \centering
  \includegraphics[trim=1.3cm 0.6cm 0cm 0cm, clip,scale=1.05]{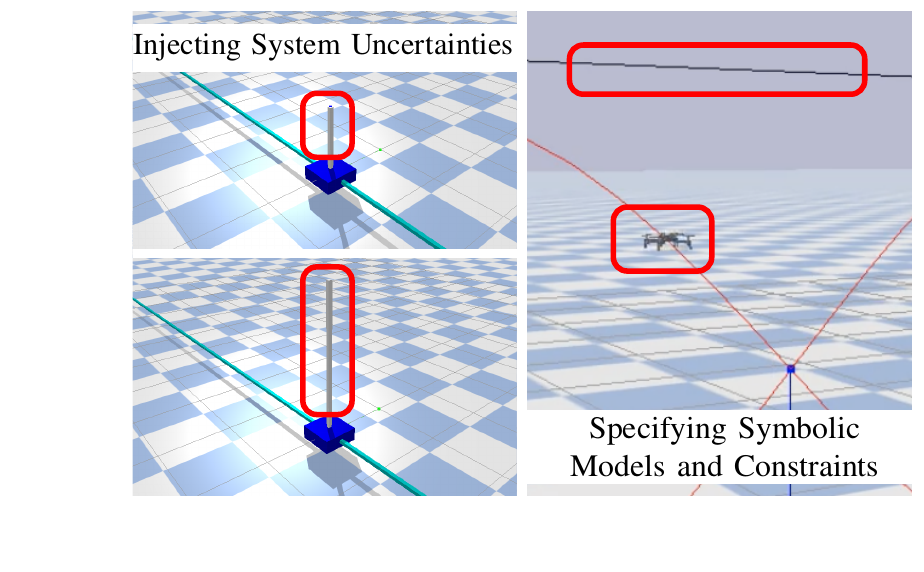}
  \caption{
  \texttt{\scriptsize safe-control-gym} includes symbolic dynamics and constraints, and allows the injection of disturbances on inputs, states, and inertial properties to support the development of safe control algorithms for both stabilization and trajectory tracking tasks.
}
  \label{fig:front-page}
\end{figure}

Our work was motivated by the lack of open-source, re-usable tools for the comparison of learning-based control research and RL (including methods that leverage prior knowledge)~\cite{DSL2021}.
While we acknowledge the importance of eventual real-robot experiments, here, we focus on simulation as a way to lower the barrier of entry and appeal to a larger fraction of the control and RL communities.
To develop safe learning-based robot control, we need a simulation environment that can \emph{(i)} support model-based approaches, \emph{(ii)} express safety constraints, and \emph{(iii)} capture real-world non-idealities (such as uncertain physical properties and imperfect state estimation).
Our ambition is for this software  to bring closer, support, and speed up the work of control and RL researchers, allowing them to easily compare results.
We strive for simple, modular, and reusable code which leverages two open-source tools popular with each of the two communities: PyBullet's physics engine~\cite{coumans2021} and CasADi's symbolic framework~\cite{Andersson2019}.

\begin{table*}[h!]
    \centering
    \caption{
    Feature comparison of \texttt{safe-control-gym} and other safety-focussed reinforcement learning environments
}
    \scalebox{0.94}{
\begin{tabular}{ c  c  >{\centering}p{1.3cm} >{\centering}p{1.5cm}  >{\centering}p{1.5cm}  >{\centering}p{1.6cm}  c  >{\centering}p{1.5cm}  c  c }
    \toprule
& Physics
    & Rendering
    & \multirow{2}{*}{Robots}
    & \multirow{2}{*}{Tasks}
    & Uncertain
    & \multirow{2}{*}{Constraints}
    & \multirow{2}{*}{Disturbances}
    & Gym
    & Symbolic \\
& Engine
    & Engine
    & 
    & 
    & Conditions
    & 
    &
    & API
    & API
    \\

    \cmidrule(lr){1-1}
    \cmidrule(lr){2-3}
    \cmidrule(lr){4-5}
    \cmidrule(lr){6-8}
    \cmidrule(lr){9-10}

    {\scriptsize \textbf{\texttt{ safe-control-gym}}}
    & {\scriptsize Bullet}
    & {\scriptsize TinyRenderer, OpenGL} & {\scriptsize Cart-pole, Quadrotor}
    & {\scriptsize Stabilization, Traj. Track.}
    & \cellcolor{black!10} {\scriptsize Inertial Param., Initial State}
    & \cellcolor{black!10} {\scriptsize State, Input}
    & \cellcolor{black!10} {\scriptsize State, Input, Dynamics}
    & \cellcolor{black!10} {\scriptsize Yes}
    & \cellcolor{black!10} {\scriptsize Yes}
    \\
    
    \cmidrule(lr){2-10}

    {\scriptsize \texttt{safety-gym}~\cite{ray2019a}}
& {\scriptsize MuJoCo}
    & {\scriptsize OpenGL}
    & {\scriptsize Point, Car, Quadruped}
    & {\scriptsize Goal, Push, Button} & \cellcolor{black!10} {\scriptsize Initial State}
    & \cellcolor{black!10} {\scriptsize State}
    & \cellcolor{black!10} {\scriptsize Adversaries}
    & \cellcolor{black!10} {\scriptsize Yes}
    & {\scriptsize No}
    \\

    \cmidrule(lr){2-10}
    
    {\scriptsize \texttt{realworldrl-suite}~\cite{dulacarnold2020a}}
& {\scriptsize MuJoCo}
    & {\scriptsize OpenGL}
    & {\scriptsize Humanoid, Cart-pole, etc.} & {\scriptsize \textcolor{black}{Swing-up, Walk, etc.}} & \cellcolor{black!10} {\scriptsize Inertial Param., Initial State}
    & \cellcolor{black!10} {\scriptsize State}
    & \cellcolor{black!10} {\scriptsize State, Input, Reward}
    & {\scriptsize No}
    & {\scriptsize No}
    \\
    
    \cmidrule(lr){2-10}

    {\scriptsize \texttt{ ai-safety-gridworlds}~\cite{leike2017a}}
& {\scriptsize n/a}
    & {\scriptsize Terminal}
    & {\scriptsize n/a}
    & {\scriptsize Grid Navigation}
    & \cellcolor{black!10} {\scriptsize Initial State, Reward}
    & \cellcolor{black!10} {\scriptsize State}
    & \cellcolor{black!10} {\scriptsize Dynamics, Adversaries}
    & {\scriptsize No}
    & {\scriptsize No}
    \\

    \bottomrule
\end{tabular}
}     \label{tab:comparison}
    \vspace{-1em}
\end{table*}

The contributions and features of the simulation environments in \texttt{\small safe-control-gym}(Figure~\ref{fig:front-page}) can be summarized as follows:
\begin{itemize}
    \item we provide open-source simulation environments with a novel, augmented \emph{Gym} API, including symbolic dynamics models and trajectory generation,
    designed to seamlessly interface with both RL and control approaches;
    \item \texttt{\small safe-control-gym} allows for constraint specification and disturbance injection onto a robot's inputs, states, and inertial properties through a portable configuration system. This is crucial to simplify the development and comparison of safe learning-based control approaches;
    \item finally, our codebase includes open-source implementations of several baselines from traditional control, RL, and learning-based control, which we use in~\cite{DSL2021} and Section~\ref{sec:results} to demonstrate how \texttt{\small safe-control-gym} supports quantitative comparisons across fields.
\end{itemize}

\revision{
As pointed out in \cite{Recht2019,DSL2021}, despite the undeniable similarities in their setup, there still exist terminology gaps and disconnects in how optimal control and reinforcement learning research address safe robot control.
In~\cite{DSL2021}, as we reviewed the last half-decade of safe robot control research, we observed significant differences in the use and reliance on prior models and assumptions.
We also found a distinct lack of open-source simulations and control implementations (Figure 5 of~\cite{DSL2021}), which are essential for reproducibility and comparisons across fields and methodologies.
With this work, we intend to make it easier for both RL and control researchers to
\emph{(i)} publish their results based on open-source simulations,
\emph{(ii)} compare against both RL and traditional control baselines, 
and \emph{(iii)} quantify safety against shared, reusable sets of constraints and/or disturbances.
}
 \section{Related Work}
\label{sec:related}

Simulation environments such as OpenAI's \texttt{\small gym}~\cite{brockman2016a} and DeepMind's \texttt{\small dm\_control}
have been proposed as a way to standardize the development of RL algorithms
\revision{and have been used, for example, in evaluation tool-kits for deep and model-based RL agents~\cite{garage,wang2019a}.}
However, recent work~\cite{drlthatmatters}
has highlighted that, even using these tools, RL research is often difficult to reproduce as it might hinge on the choice of hyper-parameter values or random seeds.

In robotics research, physics-based simulators---such as Google's Brax~\cite{freeman2021brax} and Nvidia's Isaac Gym~\cite{makoviychuk2021isaac}---are an increasingly popular tool to effectively collect large data sets that approximate complex dynamical systems~\cite{collins2021a}. While MuJoCo
has been the dominant force behind many of the physics-based RL environments, it is not yet a fully open-source project.
In this work, we leverage the Python bindings of the open-source C++ Bullet Physics engine~\cite{coumans2021} instead, which currently powers several \revision{re-implementations of the original MuJoCo tasks~\cite{pybulletgym}} as well as additional robotic simulations, including quadrotors~\cite{panerati2021learning} and quadrupeds. 

Some aspects of safety have been investigated by existing simulation environments. 
DeepMind's \texttt{\small ai-safety-gridworlds}~\cite{leike2017a}, is a set of RL environments meant to assess the safety properties
of intelligent agents by considering distributional
shifts, robustness to adversaries, and safe exploration.
However, it is not specific to real-world robotics  systems, as these environments focus on grid worlds.
OpenAI's \texttt{\small safety-gym}~\cite{ray2019a} and Google's \texttt{\small realworldrl\_suite}~\cite{dulacarnold2020a} both augment typical RL environments with constraint evaluation. They include a handful of robotic platforms such as two-wheeled robots and quadrupeds.
Similarly to our work, \texttt{\small realworldrl\_suite} also includes perturbations in actions, observations, and physical quantities.
However, unlike our work, the environments in \cite{ray2019a,dulacarnold2020a} lack the support for a symbolic framework to express \emph{a priori} knowledge of a system's dynamics or its constraints.

As our \texttt{\small safe-control-gym} includes a \emph{Gym}-style quadrotor environment, it is important to clarify that we focus on safe, low-level control rather than vision-based applications (as AirSim~\cite{shah2018} or Flightmare~\cite{song2020}) or multi-agent coordination (as \texttt{\small gym-pybullet-drones}~\cite{panerati2021learning}).

Our work advances the state-of-the-art in safety-focused simulation environments (summarized in Table~\ref{tab:comparison}) by enabling the definition of: 
\emph{(i)} symbolic models of the dynamics, cost, and constraints of an RL environment to support traditional control and model-based approaches;
\emph{(ii)} customizable, portable, and reusable constraints and physics disturbances to facilitate comparisons and enhance reproducibility;
and \emph{(iii)} several control and learning-based control baselines in addition to the typical RL baselines.
\revision{
These new features, paired with full retro-compatibility with \emph{Gym}'s original API, make \texttt{\small safe-control-gym} an ideal playground to develop and compare RL and learning-based robot control.
}

\begin{figure*}[t]
  \centering
  \hspace{-2em} \includegraphics[]{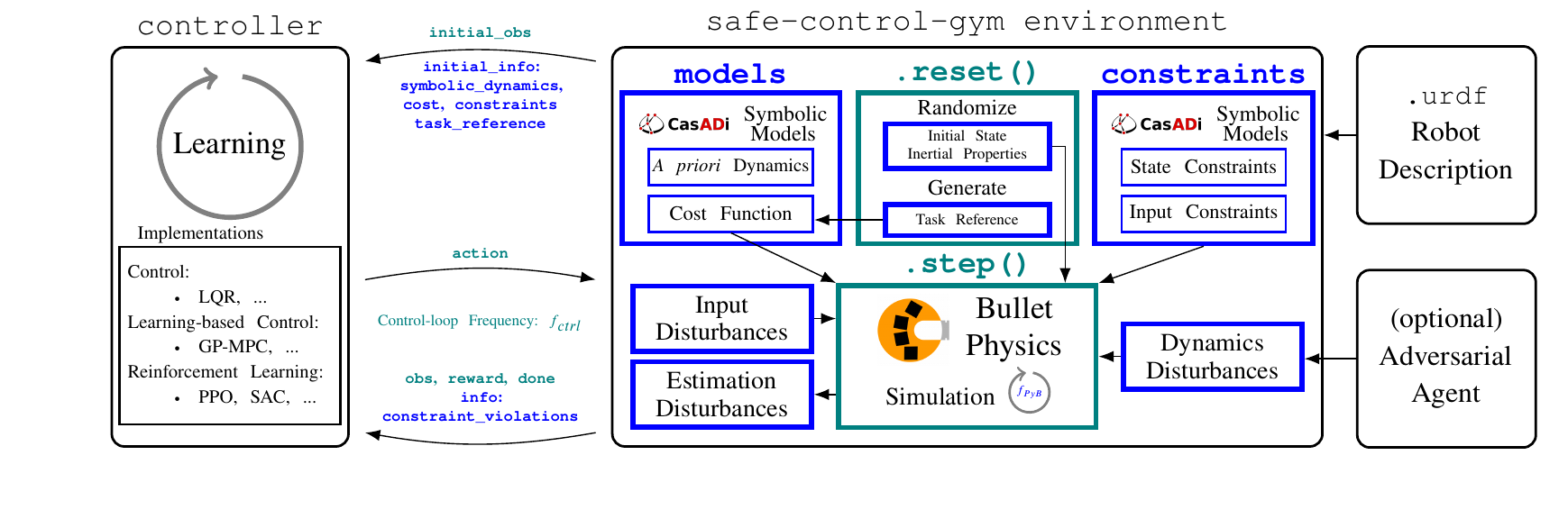}
  \vspace{-2em}
  \caption{
  Block diagram of \texttt{safe-control-gym}'s Python architecture,  highlighting the backward compatibility with OpenAI's \emph{Gym} API (teal components) and the proposed extensions (blue components) for the development of safe learning-based control and reinforcement learning.
  }
  \label{fig:block-diagram}
  \vspace{-1em}
\end{figure*}

 \section{Environments}
\label{sec:environments}

\begin{figure}[t]
  \centering
  \includegraphics[trim=0.4cm 0.35cm 0cm 0cm, clip]{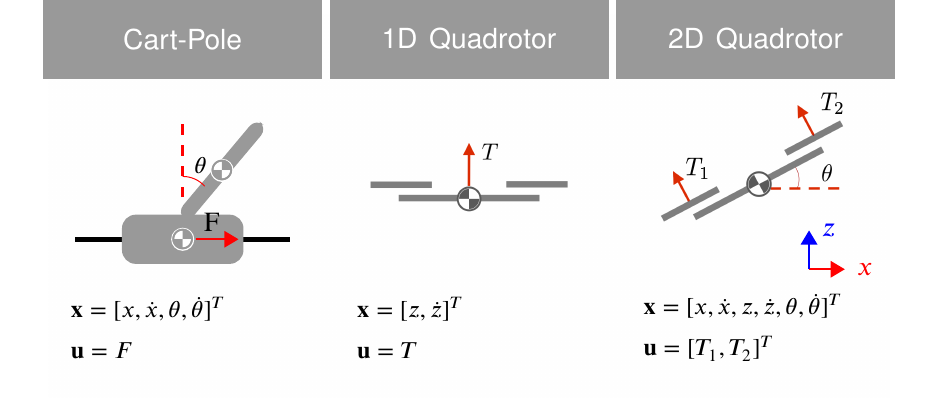}
  \caption{Schematics, state and input vectors of the cart-pole, and the 1D and 2D quadrotor environments in \texttt{\scriptsize safe-control-gym}. }
  \label{fig:envs}
\end{figure}

Our open-source suite \texttt{\small safe-control-gym} comprises three dynamical systems: the cart-pole, and the 1D and 2D quadrotors.
We consider two  control tasks: stabilization and trajectory tracking.
The software package can be downloaded and installed as \revision{(see \href{https://github.com/utiasDSL/safe-control-gym\#install-on-ubuntumacos}{\texttt{README.md}} for troubleshooting)}:
\begin{lstlisting}[firstnumber=1,language=Python,
    numbers=none,
label = {alg:install},]
$ git clone -b submission \
> git@github.com:utiasDSL/safe-control-gym.git
$ cd safe-control-gym/
$ pip3 install -e .
\end{lstlisting}

As advised in~\cite{how2015}, ``benchmark problems should be complex enough to highlight issues in controller design [...] but simple enough to provide easily understood comparisons.''.
All three systems in \texttt{\small safe-control-gym} are unstable.
We included the cart-pole, since it has been widely adopted to showcase traditional control as well as RL since the mid-80s~\cite{barto1983}.
The 1D quadrotor is a linear dynamic system, the 2D quadrotor is nonlinear, and the cart-pole is non-minimum phase.
The 1D quadrotor is a simple double-integrator system, which can also be used for didactic purposes.

\subsection{Cart-Pole System}
\label{subsec:cart-pole}

An overview of the cart-pole system is given in Figure~\ref{fig:envs}: a cart with mass $m_c$ connects \emph{via} a prismatic joint to a 1D track; a pole of mass $m_p$ and length $2l$ is hinged to the cart.
The state vector for the cart-pole is $ \mathbf{x} = [ x, \dot{x}, \theta, \dot{\theta}]^T$, where $x$ is the horizontal position of the cart, $\dot{x}$ is the velocity of the cart, $\theta$ is the angle of the pole with respect to vertical, and $\dot{\theta}$ is the angular velocity of the pole.
The input to the system is a force $\sysinput = F$ applied to the center of mass (COM) of the cart. 
In the frictionless case, the equations of motion for the cart-pole system are given in~\cite{florian2007correct} as:
\begin{subequations}
\begin{align}
    \ddot{x} &= 
    \frac{
        F + m_p l ( \dot{\theta}^2 \sin \theta - \ddot{\theta} \cos \theta )
    }{
        m_c + m_p
    }
    \label{eq:carp-pole-x-ddot}\\
        \ddot{\theta} &= 
    \frac{
        g \sin \theta + \cos \theta
               \left( \frac{-F - m_p l \dot{\theta}^2 \sin \theta }
                           {m_c + m_p}
               \right)
    }{
        l \left( \frac{4}{3}
                - \frac{m_p \cos^2 \theta}
                       {m_c + m_p}
           \right)
    }
    \label{eq:cart-pole-theta-ddot},
    \end{align}
    \label{eq:cart-pole}\end{subequations}
where
$g$ is the acceleration due to gravity.

\subsection{1D and 2D Quadrotor Systems}
\label{subsec:quad}

The second and third robotic systems in \texttt{\small safe-control-gym} are the 1D and the 2D quadrotor.
These correspond to the cases in which the movement of a quadrotor is constrained to the 1D motion in the vertical $z$-direction and the 2D motion in the $xz$-plane, respectively (Figure~\ref{fig:envs}).
For a physical quadrotor, these motions can be achieved by controlling the four motor thrusts of the quadrotor to balance out the force and torque along the redundant dimensions.

In the 1D quadrotor case, the state of the system is $\sysstate = [z, \dot{z}]^T$, where $z$ and $\dot{z}$ are the vertical position and velocity of the COM of the quadrotor. The input to the system is the overall thrust $\sysinput = T$ generated by the motors of the quadrotor. 
The equation of motion for the 1D quadrotor system is:
\begin{equation}
    \ddot{z} = T / m - g,
    \label{eq:1d-quad}
\end{equation}
where $m$ is the mass of the quadrotor and $g$ is the acceleration due to gravity. 

In the 2D quadrotor case, the state of the system is $\sysstate = [x, \dot{x}, z, \dot{z}, \theta, \dot{\theta}]^T$, where $(x,z)$ and $(\dot{x}, \dot{z})$ are the translation position and velocity of the COM of the quadrotor in the $xz$-plane, and $\theta$ and $\dot{\theta}$ are the pitch angle and the pitch angle rate, respectively. The input of the system are the thrusts $\sysinput = [T_1, T_2]^T$ generated by two pairs of motors, one on each side of the body's $y$-axis. 
The equations of motion for the 2D quadrotor system are as follows:
\begin{subequations}
    \begin{align}
    \ddot{x} &= \sin\theta\left(T_{1} + T_{2} \right) / m\\
    \ddot{z} &= \cos\theta\left(T_{1} + T_{2} \right) / m - g\\
    \ddot{\theta} &= \left(T_{2} - T_{1} \right) d/ I_{yy},
    \end{align}
    \label{eq:2d-quad}\end{subequations}
where $m$ is the mass of the quadrotor, $g$ is the acceleration due to gravity, $d = l/\sqrt{2}$ is the effective moment arm (with $l$ being the arm length of the quadrotor, i.e., the distance from each motor pair to the COM), and $I_{yy}$ is the moment of inertia about the $y$-axis.

\subsection{Stabilization and Trajectory Tracking Tasks}
\label{subsec:cost}

For all three systems detailed in Sections~\ref{subsec:cart-pole} and~\ref{subsec:quad}, our framework currently supports two control tasks:
stabilization
and trajectory tracking.
For stabilization, \texttt{\small safe-control-gym} provides an equilibrium pair for the system, $\textbf{x}^{\text{ref}}$, $\textbf{u}^{\text{ref}}$.
For trajectory tracking, we include a trajectory generation module capable of generating circular, sinusoidal, lemniscate, or square trajectories.
The module returns references $\textbf{x}^{\text{ref}}_i$, $\textbf{u}^{\text{ref}}_i \ \ \forall i \in \{0, \twodots, L\}$, where $L$ is the number of control steps in an episode.
To run a quadrotor example, tracking different trajectories with PID control, try:
\begin{lstlisting}[firstnumber=1,language=Python,
    numbers=none,
label = {alg:install},]
$ cd safe-control-gym/examples/
$ python3 tracking.py --overrides ./tracking.yaml
\end{lstlisting}

For control-based approaches, on both stabilization and trajectory tracking tasks, \texttt{\small safe-control-gym} \revision{allows the computation of} a quadratic cost $J^{Q}$ of the form:
\begin{multline}
    J^{Q} = \frac{1}{2} \textstyle\sum_{i=0}^L
    (\mathbf{x}_i - \mathbf{x}_i^{\text{ref}})^T \mathbf{Q} (\mathbf{x}_i - \mathbf{x}_i^{\text{ref}})\\ 
    + \frac{1}{2} \textstyle\sum_{i=0}^{L-1}
    (\mathbf{u}_i - \mathbf{u}_i^{\text{ref}})^T \mathbf{R} (\mathbf{u}_i - \mathbf{u}_i^{\text{ref}}),
    \label{eq:cart-pole-stab-cost}
\end{multline}
where $\mathbf{Q}$, $\mathbf{R}$ are parameters of the cost function (note that, for the stabilization task, $\textbf{x}^{\text{ref}}_i$, $\textbf{u}^{\text{ref}}_i$ are not time-varying).

In RL, an agent's performance is expressed by the total collected reward $J^{R}$ and we use the negated quadratic cost to express it, that is $J^{R} = -J^{Q}$.
The RL state $\mathbf{x}_i$ is also augmented with the target position within the trajectory $\mathbf{x}_i^{\text{ref}}$. To enable further comparisons, we also implement the traditional reward function for cart-pole stabilization found in~\cite{barto1983,brockman2016a}.
This is an instantaneous reward  of 1 for each discrete time step $i$ in which the pole is upright, i.e., $|\theta| \leq \theta_{\text{max}}$. Episodes are terminated on step $D$, when $|\theta|$ exceeds threshold $\theta_{\text{max}}$ or $D = L$, thus $J^{R} = D$.

\subsection{\texttt{\small safe-control-gym} Extended API}
\label{subsec:api}

To provide native support to open-source RL libraries,
\texttt{\small safe-control-gym} adopts OpenAI \emph{Gym}'s interface.
However, to the best of our knowledge, we are the first to extend this API with the ability to provide a learning agent with \emph{a priori} knowledge of the dynamical system.
This is of crucial importance to also support the development of, and comparison with, learning-based control approaches, which typically leverage insights about the physics of a robotic system.
The proposed gym environment allows for this prior information to be integrated into the learning process.
An overview of \texttt{\small safe-control-gym}'s modules
is presented in Figure~\ref{fig:block-diagram}.
Our benchmark suite can be used, for example, to answer the question of how data-efficiency, which is imperative for fast robot learning, is impacted by model-free RL approaches that do not exploit prior knowledge (\textcolor{black}{see Section~\ref{sec:results}}).
To run one of \texttt{\small safe-control-gym}'s environments, in headless mode, with terminal printouts from the original Gym API \textcolor{black}{(in cyan)} and our new API \textcolor{black}{(in purple)}, try:
\begin{lstlisting}[firstnumber=1,language=Python,
    numbers=none,
label = {alg:install},]
$ cd safe-control-gym/examples/
$ python3 verbose_api.py --system cartpole \
> --overrides verbose_api.yaml
\end{lstlisting}

The next four subsections detail the three core features of our simulations environments---\emph{a priori} symbolic models, constraint specification, and disturbance injection---and the configuration system used for experiments' reproducibility.

\subsubsection{Symbolic Models}

We use CasADi~\cite{Andersson2019}, an open-source symbolic framework for nonlinear optimization and algorithmic differentiation, to include symbolic models of \emph{(i)} the systems' \emph{a priori} dynamics (see~Sections~\ref{subsec:cart-pole} and~\ref{subsec:quad})
as well as \emph{(ii)} the quadratic cost function from Section~\ref{subsec:cost},
and \emph{(iii)} optional constraints (see Section~\ref{subsubsec:constr}).
As shown by the printouts of the snippet above, these models,
together with the initial state of the system $\mathbf{x}_0$ and task references $\mathbf{x}^{\text{ref}}$, $\mathbf{u}^{\text{ref}}$,
are returned by our API in a {\small \texttt{reset\_info}} dictionary at each reset of an environment.

\subsubsection{Constraints}
\label{subsubsec:constr}

The ability to specify, evaluate, and enforce one or more constraints $c^j \ \ j \in \{0, 1, 2, \twodots\}$ on state $\mathbf{x}$ and input $\mathbf{u}$:
\begin{equation}
    c^j (\mathbf{x}_i, \mathbf{u}_i) \leq 0   \quad \quad \quad \forall i \in \{0, \twodots, L\},
    \label{eq:tbd}
\end{equation}
is essential for safe robot control.
While previous RL environments including state constraints exist~\cite{ray2019a,dulacarnold2020a}, our implementation is the first to also provide their symbolic representation and the ability to add bespoke ones while creating an environment (see Section~\ref{subsubsec:config}).
Our current implementation includes default input and state constraints and supports user-specified ones in multiple forms (linear, bounded, quadratic) on either the system's state, input, or both.
Constraint evaluations are included in the {\small \texttt{info}} dictionary returned at each environment's step (see Figure~\ref{fig:block-diagram}).
\revision{With this constraint API, \texttt{\small safe-control-gym} can accommodate the different conventions and usage of constraints in safe control and safe RL, providing the necessary support to fairly evaluate and compare their safety properties.}

\subsubsection{Disturbances}
\label{subsubsec:dist}

In developing safe control approaches, we are often confronted with the fact that models like the ones in Sections~\ref{subsec:cart-pole}
and~\ref{subsec:quad} are not a complete or fully truthful representation of the system under test.
\texttt{\small safe-control-gym} provides several ways to implement non-idealities that mimic real-life robots, \revision{covering many of the typical test scenarios in safe learning-based control as well as robust RL research,} including:
\begin{itemize}
	\item randomization (from a given probability distribution) of the initial state of the system, $\mathbf{x}_0$;
	\item randomization (from given probability distributions) of the inertial parameters, that is, $m_c$, $m_p$, $l$ for the cart-pole and $m$, $I_{yy}$ for the quadrotor; 
	\item disturbances (in the form of white noise, step, or impulse) applied to the action input $\mathbf{u}$ sent from the controller to the robot;
	\item disturbances (in the form of white noise, step, or impulse) applied to the observations of the state $\mathbf{x}$ returned by an environment to the controller;
	\item \revision{external dynamics disturbances, additional forces applied to a robot using PyBullet APIs mimicking, for example, wind and other aerodynamic effects or set deterministically (from another agent outside the environment) to implement robust adversarial training schemes such as the one proposed in~\cite{pinto2017robust}.}
\end{itemize}

\subsubsection{Configuration System}
\label{subsubsec:config}

To facilitate the reproducibility and portability of experiments with identical environment setups, \texttt{\small safe-control-gym} includes a YAML configuration system that supports all of the features discussed in Sections~\ref{subsec:cost} and~\ref{subsec:api} \revision{(see ``Configuration'' in \href{https://github.com/utiasDSL/safe-control-gym\#configuration}{\texttt{README.md}})}.

\subsection{Computational Performance}

The ability to collect large experimental datasets or generate representative simulated ones is one of the bottlenecks of learning-based robotics. With this in mind, we assessed the computational performance of \texttt{\small safe-control-gym} on a system with a 2.30GHz Quad-Core i7-1068NG7 CPU, 32GB 3733MHz LPDDR4X of memory, and running Python 3.8 under macOS 12.
\revision{In the \href{https://github.com/utiasDSL/safe-control-gym\#performance}{repository's documentation}, we summarize}
the simulation speed-ups (with respect to the wall-clock) obtained for the cart-pole and 2D quadrotor environments, in headless mode or using the GUI, with or without constraint evaluation, and different choices of control and physics integration frequencies.
In headless mode, a single instance of \texttt{\small safe-control-gym} allows to collect data 10 to 20 times faster than in real life, with fine-grained physics stepped by PyBullet at 1000Hz.

 \section{Control Algorithms}
\label{sec:algorithms}

The codebase of \texttt{\small safe-control-gym} also comprises several implementations of core control approaches, from traditional and learning-based control, to safety-certified control and safe reinforcement learning~\cite{DSL2021}.

\subsection{Control and Safe Control Baselines}

Our benchmark suite includes, as baselines, standard state-feedback controllers such as the linear quadratic regulator (LQR) and iterative LQR (iLQR)\cite{buchli2017optimal}.
LQR assumes linear system dynamics (as in~\eqref{eq:1d-quad}) and quadratic cost (as in~\eqref{eq:cart-pole-stab-cost}).
For nonlinear systems (e.g., the 2D quadrotor in~\eqref{eq:2d-quad} and cart-pole in~\eqref{eq:cart-pole}), LQR uses local linear approximations of the nonlinear dynamics.
Unlike LQR, iLQR iteratively improves the performance 
by finding better local approximations of the cost function~\eqref{eq:cart-pole-stab-cost} and system dynamics using the state and input trajectories from the previous iteration.
Our \texttt{\small safe-control-gym} environments expose the symbolic models of \textit{a priori} dynamics and cost function, facilitating the computation of dynamics' Jacobians and the Jacobians and Hessians of the cost function.
We include LQR and iLQR to show how our benchmark supports model-based approaches and
the symbolic expressions of the first-order and second-order terms included in each environment can be equivalently leveraged by other model-based methods.

We also implemented two predictive control baselines: linear model predictive control (LMPC) and nonlinear model predictive control (NMPC)~\cite{ralings2020mpc}.
At every control step, model predictive control (MPC) solves a constrained optimization problem to find a control input sequence, over a finite horizon, that minimizes the cost 
subject to the system's predicted dynamics and possibly input and state constraints.
Then, only the first optimal control input from the sequence is applied.
While NMPC uses the nonlinear system model,
LMPC uses the linearized approximation to predict the evolution of the system, sacrificing prediction accuracy for computational efficiency.
In our codebase, CasADi's \texttt{\small opti} framework~\cite{Andersson2019} is used to formulate the optimization problem.
\texttt{\small safe-control-gym} provides all the system's components required by MPC (\textit{a priori} dynamics, constraints, cost function) as CasADi models, see Section~\ref{subsec:api} for details.

\subsection{Reinforcement Learning Baselines}

As \texttt{\small safe-control-gym} extends the original \emph{Gym} API, any compatible RL algorithm can directly be applied to our environments.
In our codebase, we include two of the most well-known RL baselines:
Proximal Policy Optimization (PPO)~\cite{schulman2017proximal} and Soft Actor-Critic (SAC)~\cite{sac}. These are model-free approaches that map state measurements to control inputs without leveraging a dynamics model and using neural network (NN)-based policies.
Both PPO and SAC have been shown to work on a wide range of simulated robotics tasks, some of which involve hybrid and nonlinear dynamics. We adapt their implementations from \texttt{\small stable-baselines3}~\cite{stable-baselines3} and OpenAI's \emph{Spinning Up},
with a few modifications to also support our suite's configuration systems.
PPO and SAC are not natively safety-aware approaches and do not guarantee constraint satisfaction nor robustness beyond the generalization properties of NNs.

\subsection{Safe Learning-Based Control}

Safe learning-based control approaches improve a robot's performance using past data to improve the estimate of a system's true dynamics.
These approaches typically provide guarantees on stability and/or constraint satisfaction. 
One of these approaches, included in \texttt{\small safe-control-gym}, is GP-MPC~\cite{Hewing2019}. 
This method models uncertain dynamics with a Gaussian process (GP), which is used to better predict the future evolution of the system as well as tighten constraints based on the confidence of the dynamics along the prediction horizon. 
GP-MPC has been demonstrated on off-road ground robots~\cite{Hewing2019}.
Our implementation leverages the NMPC controller, the environments' symbolic \emph{a priori} model, and uses \texttt{\small gpytorch}~\cite{gardner2021gpytorch} for the GP modelling and optimization of the uncertain dynamics.
GP-MPC can accommodate both environment and controller-specific constraints.

\subsection{Safe and Robust Reinforcement Learning}

Building upon the RL baselines, we implemented three safe RL approaches that address the problems of constraint satisfaction and robust generalization against disturbances from fixed distributions.
The \emph{safety layer}-based approach in~\cite{dalal2018a} pre-trains NN models to approximate linearized state constraints.
These learned constraints are then used to filter potentially unsafe inputs from an RL controller \emph{via} least-squares projection (also see Section~\ref{subsec:safety_certification}). \revision{Once pre-trained (without the need for a prior model), the safety layer is kept fixed during an agent's learning.}
We add such a \emph{safety layer} to PPO and apply it to our benchmark tasks. Robust RL aims to learn policies that generalize across systems or tasks. We adapt two methods based on adversarial learning: RARL~\cite{pinto2017robust} and RAP~\cite{vinitsky2020robust}.
These approaches model disturbances as a learning adversary and train the policy against increasingly stronger disturbances, which create
harsher control scenarios.
The controllers in~\cite{pinto2017robust, vinitsky2020robust} demonstrated, in simulation, robustness
against parameter mismatch and specific types of disturbances.
These methods can be directly trained in \texttt{\small safe-control-gym}, by leveraging the disturbances API (see Section~\ref{subsubsec:dist}).

\begin{figure}\centering
  \includegraphics[]{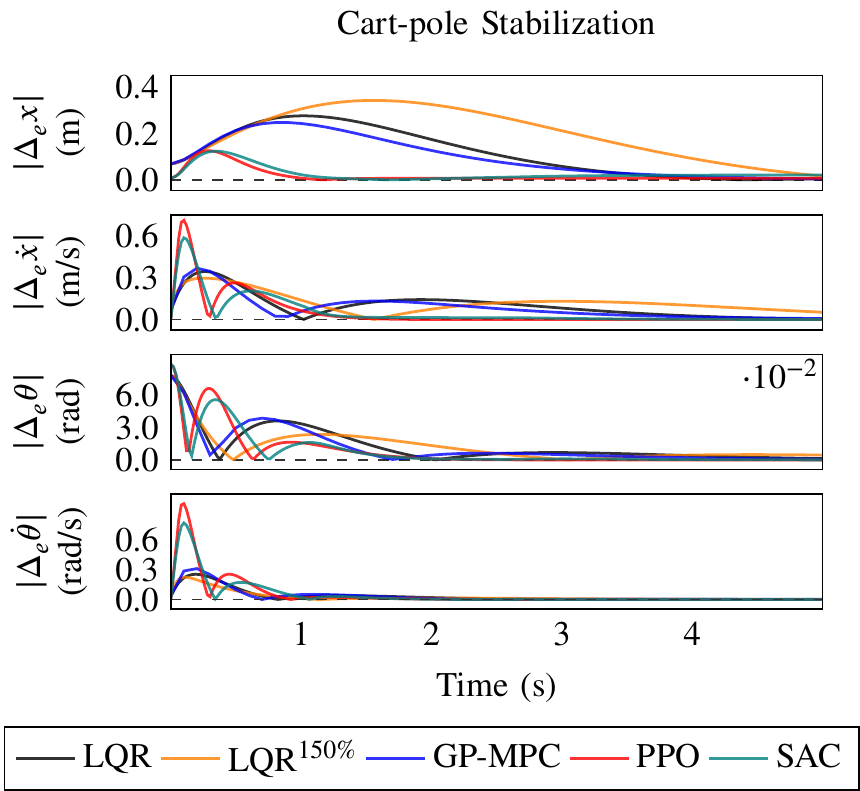}
  \caption{Control performance (absolute error w.r.t. reference $\mathbf{x}^{\text{ref}}$, $|\Delta_e|$) on the cart-pole stabilization task for different controllers and RL agents. 
  }
  \label{fig:control-cart}
\end{figure}

\subsection{Safety Certification of Learned Controllers}
\label{subsec:safety_certification}

Learned controllers lacking formal guarantees themselves can be rendered safe by augmenting them with safety filters.
These filters minimally modify unsafe control inputs, so that the applied control input maintains the system's state within a safe set.  
{\color{black}
A common safety filter that implements this idea is model predictive safety certification~(MPSC), which solves a finite-horizon constrained optimization problem with a discrete-time predictive model to prevent a learning-based controller from violating constraints~\cite{Wabersich2018a}.
In~\cite{DSL2021}, we presented an implementation of MPSC for PPO simultaneously leveraging the CasADi \textit{a priori} dynamics and constraints and the \emph{Gym} RL interface of \texttt{\small safe-control-gym}.
}

Control barrier functions (CBF) can act as safety filters for continuous-time nonlinear control-affine systems through a quadratic program (QP) with a constraint on the CBF's time derivative
~\cite{ames2019a}.
In the case of model errors, the resulting errors in the CBF's time derivative can be learned by a NN~\cite{taylor2020a}.
Learning-based CBF filters have been applied to safely control a segway~\cite{taylor2020a} \textcolor{black}{and a quadrotor~\cite{Wang2018a}}.
Again, our CBF implementation relies on the \textit{a priori} model and constraints exposed by \texttt{\small safe-control-gym}'s API.
The CBF's time derivative is also efficiently determined using CasADi.
Constraints can be handled as long as the constraint set contains the safe set defined by the CBF.

\section{Results}
\label{sec:results}

\begin{figure}\centering
  \includegraphics[]{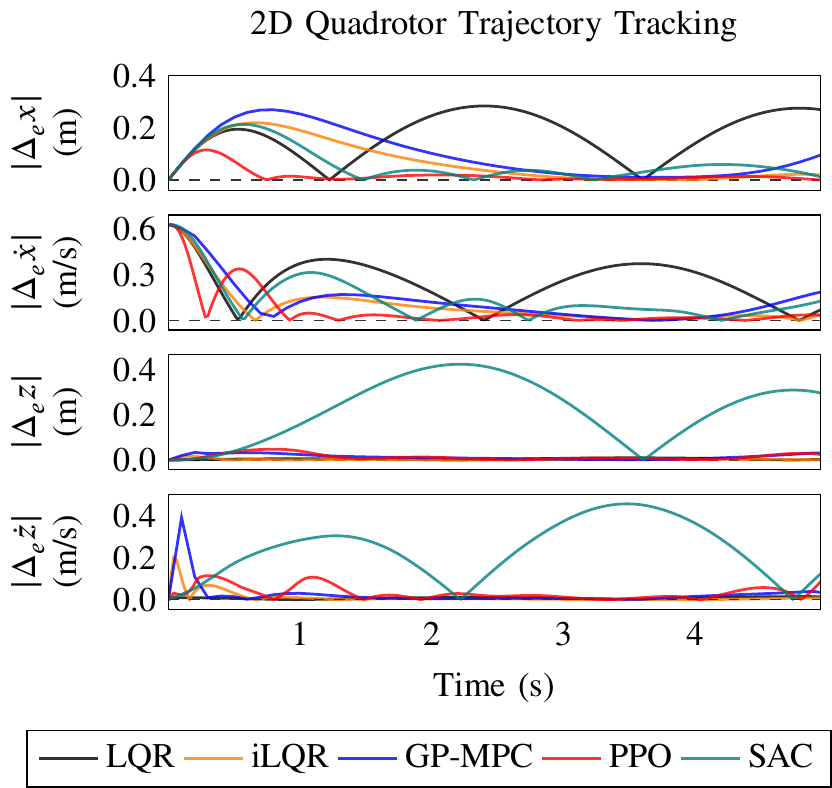}
  \caption{
Control performance (absolute error w.r.t. reference $\mathbf{x}^{\text{ref}}$, $|\Delta_e|$) on the 2D quadrotor tracking task for different controllers and RL agents.
  }
  \label{fig:control-quad}
\end{figure}

\begin{figure}\centering
  \includegraphics[]{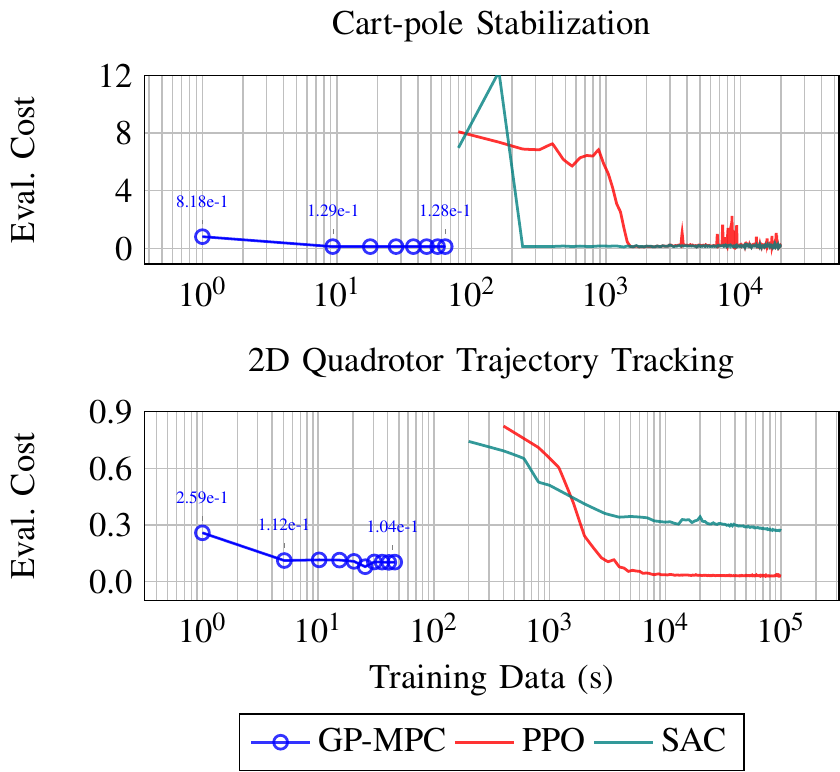}
  \caption{
  Comparison of the learning performance (on a logarithmic $x$-axis) of learning-based control (GP-MPC) and two RL agents (PPO, SAC). \revision{Training data is expressed as seconds of simulation time used for collection.} \revision{Evaluation cost is the RMSE with respect to the target state(s). In cart-pole stabilization, the RMSE uses the full state; in 2D quadrotor tracking, the RMSE only includes the $x$ and $z$ component of the state.}
  }
  \label{fig:learning}
\end{figure}

In this section, we demonstrate how \texttt{\small safe-control-gym} can be used to compare approaches from all the families of control algorithms discussed in Section~\ref{sec:algorithms}, with respect to their control performance (Figures~\ref{fig:control-cart} and \ref{fig:control-quad}), learning efficiency (Figure~\ref{fig:learning}), capability to satisfy constraints (Figure~\ref{fig:constraints}), as well as robustness against disturbances and parametric uncertainty (Figure \ref{fig:robustness}).
Our goal here is to highlight the potential for quantitative comparisons among controllers (although conditional to adequate and careful parameter tuning) as well as their qualitative features. Notably, \texttt{\small safe-control-gym} allows plotting RL and control results on a common set of axes with respect to \revision{the aforementioned} performance and safety metrics\revision{, even when different controllers self-tune for different optimization objectives}.

\subsection{Control Performance}
\label{sec:control_performance}

In Figures~\ref{fig:control-cart} and~\ref{fig:control-quad}, we show that LQR,
LQR$\mathrm{}^{150\%}$, 
GP-MPC \revision{(these latter two controllers using a prior model of the dynamics with parameters overestimated by 150\%)}, PPO, and SAC are able to stabilize the cart-pole and track the quadrotor reference trajectory (a circle with a 1 meter radius). 
For the stabilization task, GP-MPC closely matches the closed-loop trajectory of the LQR with true parameters, albeit its \textit{a priori} model was the same one given to LQR$\mathrm{}^{150\%}$.
This shows how GP-MPC can overcome imperfect initial knowledge through learning.
Both PPO and SAC yield substantially different closed-loop trajectories when compared to LQR and GP-MPC. 
\revision{Besides the fact that they have distinct learning procedures and parameterization, this is also due to the difference in their optimization objectives, and differences in prior information the algorithms have access to.}
Tracking the quadrotor sinusoidal trajectories (Figure~\ref{fig:control-quad}) results in low-frequency oscillations for PPO and SAC, \revision{a consequence possibly from learning with stochastic data and the lack of explicit long-term planning}. The LQR results in relatively large tracking errors due to the inaccuracy of the linearized dynamics model in approximating the nonlinear quadrotor system, especially for aggressive maneuvers.

\subsection{Learning Performance and Data Efficiency}

Figure \ref{fig:learning} shows how much data GP-MPC, PPO, and SAC require to achieve comparable trajectory root-mean-squared error \revision{(RMSE)}, a common metric for comparing controller performance.
\revision{The values reported on the $x$-axis (training data) are measured in seconds of simulation time used for learning, derived as the number of elapsed environment simulation steps and PyBullet's simulation frequency.}
This plot showcases the type of interdisciplinary comparisons uniquely enabled by \texttt{\small safe-control-gym}.
In both plots, the untrained GP-MPC displays a performance that is only matched after around $10^3$ seconds of simulated data
by the RL approaches.
GP-MPC converges to its optimal performance with roughly one tenth (or less) of the data. 
This highlights how learning-based control approaches
are orders of magnitude more data-efficient than model-free RL. 
However, this is largely the result of knowing a reasonable \textit{a priori} model.
The evaluation costs of PPO and SAC exhibit large oscillations and learning instability in the early stage, not uncommon in deep RL~\cite{drlthatmatters}. Once converged SAC and PPO reach a performance comparable to GP-MPC in the stabilization task. PPO can match 
GP-MPC on the tracking task \revision{albeit using much more data. Although optimal hyper-parameter tuning is beyond the scope of this work, it should be noted that, for example, PPO's learning stability is improved by a larger batch size, especially in the quadrotor tracking task.} 

\begin{figure}\centering
  \includegraphics[]{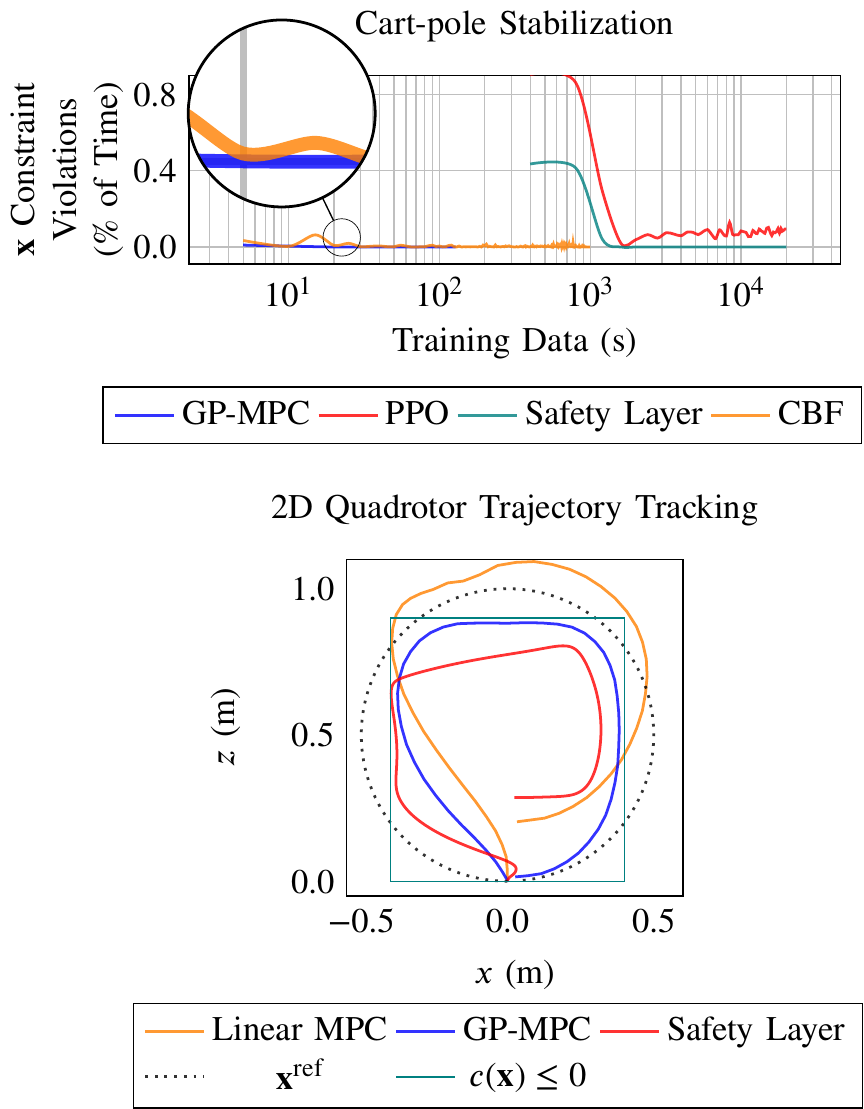}
  \caption{
  In the top plot, fraction of time spent incurring a constraint violation by learning-based control (GP-MPC), vanilla RL (PPO), safety-augmented RL (Safety Layer), and safety-certified control (CBF); in the bottom plot, trajectories for an ``impossible'' tracking task (with constraints narrower than the reference) for traditional control (Linear MPC), learning-based control (GP-MPC), and safety-augmented RL (PPO with Safety Layer). 
  }
  \label{fig:constraints}
\end{figure}

\subsection{Safety: Constraint Satisfaction}

In Figure~\ref{fig:constraints}, we investigate the impact of learning and the amount of training data on the constraint violations of a learning-based controller or safe RL agent. 
The top plot summarizes the data efficiency of these approaches on the cart-pole stabilization task. 
Again, leveraging \revision{the same overestimated prior model as in Section \ref{sec:control_performance}}, GP-MPC and the learning-based CBF require much fewer training examples to minimize the number of constraint violations than PPO with a \emph{safety layer}.
After training, GP-MPC, learning-based CBF, and safety layer PPO all achieve similar constraint satisfaction performance. 
Vanilla PPO also reduces the number of constraint violations but cannot match the performance of the GP-MPC and the learning-based CBF.

The bottom plot of Figure~\ref{fig:constraints} shows reduced constraint violations for GP-MPC and PPO with a \emph{safety layer} for the 2D quadrotor tracking task. 
Compared to a linear MPC with overestimated parameters, the GP-MPC meets the constraints, finding a compromise between performance and constraint satisfaction. 
PPO with a \emph{safety layer}, on the other hand, 
struggles to balance between tracking the desired trajectory and fully guaranteeing constraint satisfaction.

\begin{figure}\centering
  \includegraphics[]{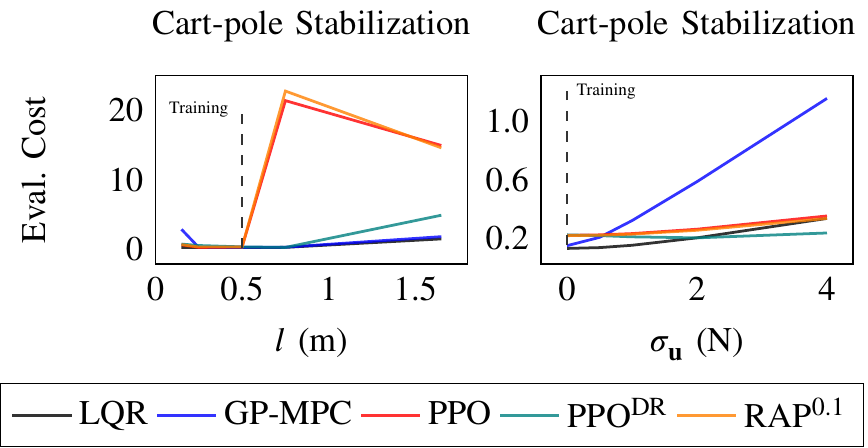}
  \caption{
  Robustness of cart-pole stabilization policies learned by traditional (LQR) and learning-based control (GP-MPC) as well as vanilla (PPO) and robust RL (PPO$^{\text{DR}}$, RAP) with domain randomization against perturbations in the length of the pole $l$ (left) and a white noise disturbance $\sim \mathcal{N}(0, \sigma_{\mathbf{u}}^2)$ applied to action input $\mathbf{u}$ (right).
}
  \label{fig:robustness}
\end{figure}

\subsection{Safety: Robustness}

\revision{Leveraging \texttt{\small safe-control-gym}'s APIs for disturbances}, Figure~\ref{fig:robustness} shows 
the robustness of controllers and RL agents against parametric uncertainty (in the pole length) and white noise (on the input) for the cart-pole stabilization task. $\mathrm{PPO}^{\text{DR}}$ is trained with pole length randomization and improves the robustness of baseline PPO. RAP is trained against adversarial input disturbances and shows robust performance to input noise, \revision{as expected from its adversarial training. However, RAP is less robust to different classes of disturbances, such as parameter mismatch.}
Model-based approaches, like LQR and GP-MPC, appear to be less affected by parameter uncertainty than model-free RL.
\revision{This could be due to the fact that a prior dynamics model, albeit inexact, helps generalizing across similar systems. However, LQR and GP-MPC are no less resilient to input noise, when compared to PPO or RAP.
GP-MPC, in particular, might be hindered by its lower control frequency (chosen to improve computational run-time), which results in accumulation of input noise that further degrades performance}.

 \section{Conclusions and Future Work}
\label{sec:conclusions}

In this letter, we introduced \texttt{\small safe-control-gym}, a benchmark suite of simulation environments to evaluate safe learning-based control.
In \texttt{\small safe-control-gym}, we combine a physics engine-based simulation with the description of the available prior knowledge and safety constraints using a symbolic framework.
By doing so, we allow the development and test of a wide range of approaches, from model-free RL to learning-based MPC.
We hope that \texttt{\small safe-control-gym} will make it easier for researchers from the RL and control communities to compare their progress, especially for the quantification of safety and robustness in robotics applications.

\balance

\balance
\end{document}